\documentclass{article}

\usepackage{microtype}
\usepackage{graphicx}
\usepackage{subfigure}
\usepackage{booktabs} 
\usepackage{multirow}
\usepackage{hyperref}

\usepackage[table,xcdraw]{xcolor}

\usepackage[accepted]{icml2024}

\usepackage{amsmath}
\usepackage{amssymb}
\usepackage{mathtools}
\usepackage{amsthm}
\usepackage{graphicx}  
\usepackage{caption}
\usepackage{subcaption}  
\definecolor{my_green}{HTML}{E5F3E0}
\definecolor{my_blue}{HTML}{DBECFC}
\definecolor{my_red}{HTML}{F0DAE3}

\usepackage[capitalize,noabbrev]{cleveref}

\theoremstyle{plain}

\theoremstyle{definition}

\theoremstyle{remark}

\usepackage[disable,textsize=tiny]{todonotes}

\icmltitlerunning{Towards Adversarially Robust Vision-Language Models}

\begin{document}

\twocolumn[
\icmltitle{Towards Adversarially Robust Vision-Language Models: \\ Insights from Design Choices and Prompt Formatting Techniques}



\icmlsetsymbol{equal}{*}

\begin{icmlauthorlist}
\icmlauthor{Rishika Bhagwatkar}{yyy,comp,cerc}
\icmlauthor{Shravan Nayak}{yyy,comp}
\icmlauthor{Reza Bayat}{yyy,comp,cerc}
\icmlauthor{Alexis Roger}{yyy,comp,cerc}
\icmlauthor{Daniel Z Kaplan}{cerc,dan}
\icmlauthor{Pouya Bashivan}{yyy,sch}
\icmlauthor{Irina Rish}{yyy,comp,cerc}
\end{icmlauthorlist}

\icmlaffiliation{yyy}{Mila – Quebec AI Institute}
\icmlaffiliation{comp}{Université de Montréal}
\icmlaffiliation{sch}{McGill University}
\icmlaffiliation{cerc}{CERC-AAI}
\icmlaffiliation{dan}{Realiz.ai}

\icmlcorrespondingauthor{Rishika Bhagwatkar}{rishika.bhagwatkar@umontreal.ca}


\vskip 0.3in
]



\printAffiliationsAndNotice{}  

\begin{abstract}
Vision-Language Models (VLMs) have witnessed a surge in both research and real-world applications. However, as they are becoming increasingly prevalent, ensuring their robustness against adversarial attacks is paramount. This work systematically investigates the impact of model design choices on the adversarial robustness of VLMs against image-based attacks. Additionally, we introduce novel, cost-effective approaches to enhance robustness through prompt formatting. By rephrasing questions and suggesting potential adversarial perturbations, we demonstrate substantial improvements in model robustness against strong image-based attacks such as Auto-PGD. Our findings provide important guidelines for developing more robust VLMs, particularly for deployment in safety-critical environments.


\end{abstract}

\section{Introduction}
\label{intro}


\begin{figure}
    \centering
    \includegraphics[width=\columnwidth]{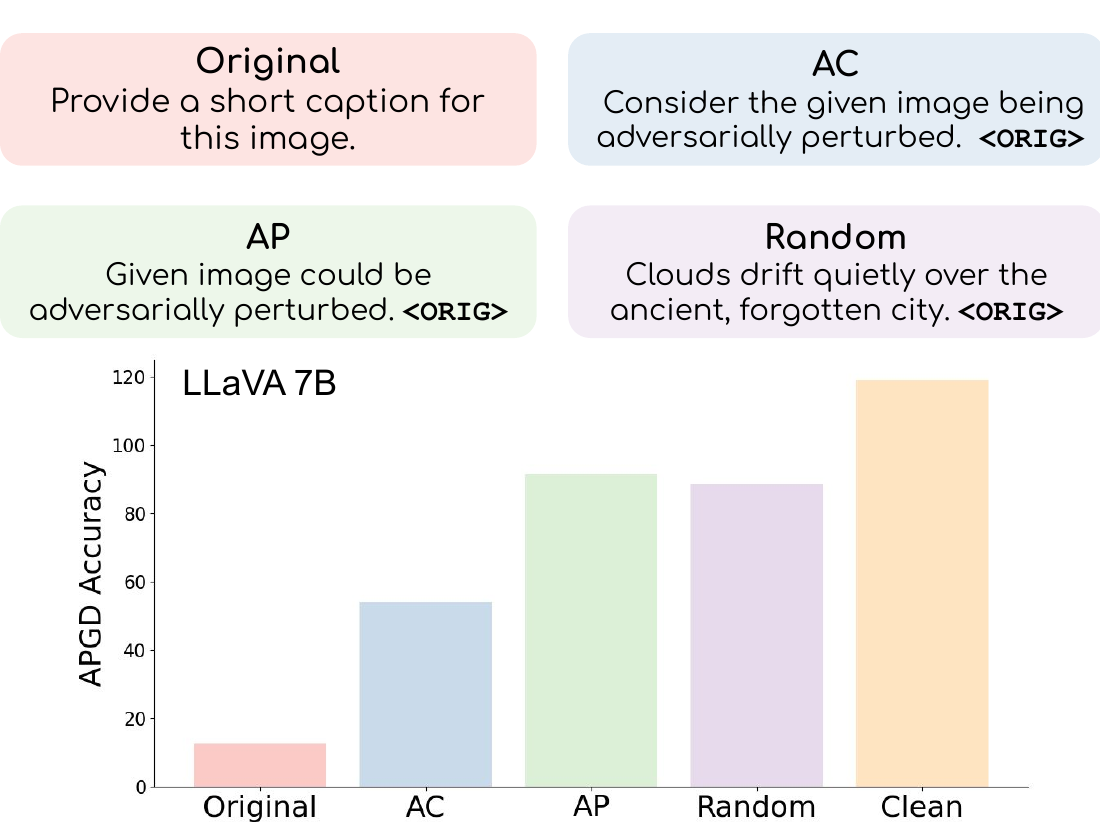}
    \caption{ Performance of LLaVA-7B on the COCO dataset when the adversarial images are given along with different types of prompts (Original, AC, AP and Random). Clean accuracy represents the model's performance on unperturbed images.\looseness-1}
    \label{fig:prompt_captioning}
\end{figure}

VLMs process images in conjunction with text prompts, enabling them to perform a wide array of tasks, such as image captioning, visual question answering (VQA), and cross-modal retrieval \cite{llava, idefics, openflamingo, clip}. While there is extensive research on advancing architecture and scaling, recent works demonstrate that VLMs are not immune to adversarial vulnerabilities — subtle, intentionally crafted perturbations to input data that can lead to significant errors in the output \cite{robustclip}. These vulnerabilities can mislead users with harmful or toxic responses, undermining the models' robustness and integrity.\looseness-1


White-box attacks are a common type of adversarial threat. These attacks assume complete access to a model’s parameters, enabling attackers to exploit specific vulnerabilities. Since many VLMs are open-source, attackers can easily analyze and exploit them. Current VLMs have several design choices, including the vision encoder, large language model (LLM), mapping network, image resolution, and the training data \cite{llava, idefics, openflamingo}. 
Despite the importance of these factors, their impact on adversarial robustness is under-explored. 
In this study, we evaluate how these design choices during VLM training influence their susceptibility to white-box adversarial attacks on the input images.

In addition to design choices, the selection and quality of prompts can significantly impact the performance and robustness of VLMs \cite{awal}. Effective prompts can enhance the models' understanding and response to inputs, affecting their robustness to adversarial attacks. Recent works have focused on adversarial training to increase robustness \cite{robustclip, mao}, but it is resource-intensive and costly, often requiring millions of samples \cite{adv_train_sota, adv_rl}. As a practical and cost-effective alternative, we investigate whether prompt formatting can enhance the adversarial robustness of VLMs. This approach explores if simple linguistic modifications can increase robustness, offering a low-cost alternative to adversarial training.
Through evaluating both design choices and prompt formatting, we aim to provide comprehensive insights into enhancing the adversarial robustness of VLMs. Our main contributions are summarized as follows:

\begin{enumerate}
    \item We provide an in-depth analysis of how various design choices of VLMs impact their robustness to white-box visual adversarial attacks.
    \item We investigate a novel approach to prompt formatting for enhancing the adversarial robustness of VLMs.
    \item To the best of our knowledge, we are the first to offer actionable insights and practical recommendations for using text prompting techniques to enhance the robustness of VLMs in deployment.
\end{enumerate}

\section{Related Works}


\textbf{Vision Language Models.} VLMs traditionally align visual tokens from the vision encoder with the linguistic space of the language model using various mapping networks, such as the Q-former in BLIP2 \cite{blip2} and the multilayer perceptron in LLaVA \cite{llava}. Recent studies investigate how choices like vision encoder type, language model, resolution of images, and training duration affect the accuracy on clean inputs \cite{prismatic}. In contrast, our study specifically aims to explore how these choices affect robust accuracy.

\textbf{Adversarial Robustness of VLMs.} 
Research into the adversarial robustness of multi-modal foundation models like BLIP2 \cite{blip2}, OpenFlamingo \cite{openflamingo}, CLIP \cite{clip}, and LLaVA \cite{llava} has highlighted their susceptibility to both targeted and untargeted visual attacks \cite{evalrobustness, eval_bb}. Studies also explore the potential of using pretrained VLMs to craft adversarial image and text perturbations that can compromise black-box models fine-tuned for various tasks \cite{eval_bb, bard}. 
Additionally, the transferability of these attacks is well-studied, with techniques developed to enhance efficacy using surrogate models \cite{vlattack}. \looseness-1

\textbf{Advancements in Defense Mechanisms.} 
Many studies focusing on the adversarial robustness of VLMs using CLIP as a vision encoder have revealed its susceptibility to adversarial attacks \cite{clip_ir_1, clip_ir_2, clip_ir_3}. To counter this, TeCoA \cite{mao} proposes adversarial fine-tuning to maintain zero-shot capabilities. Further, RobustCLIP \cite{robustclip} proposes an unsupervised method leveraging adversarial training on the ImageNet \cite{imagenet} dataset to improve robustness across vision-language tasks. Additionally, efforts to enhance robustness include prompt tuning, where one study suggests enriching prompts with contextual image-derived information for improving adversarial robustness \cite{useful_context}. Another approach optimizes prompts through adversarial fine-tuning on ImageNet with specific parameters \cite{adv-prompt}. Our research, however, focuses on analyzing the impact of prompt formatting on model performance without additional training or image-based information extraction.

\section{Experiments}

In this section, we outline the attack setups used during evaluations, the tasks assessed, and the specific models examined in our model design choice experiments.

\subsection{Attack Setup}

This work focuses on white-box gradient-based untargeted attacks on image inputs, where it is assumed that the attacker has complete knowledge of the model, including architecture and parameters. The objective in crafting adversarial samples under this scenario is to subtly perturb the input so that the model produces an incorrect output. Mathematically, it can be formulated as \(\max_{\delta} \mathcal{L}(f(x+\delta), y)\) where \(f\) is the model, \(x\) is original input, \(\delta\) is the adversarial perturbation learnt within the \(\| \delta \|_{\infty} \leq \epsilon \) constraint and \(y\) is the original label. Hence the goal is to find a perturbation \(\delta\) that maximizes the loss while respecting the perturbation bound.

Our evaluation encompasses three gradient-based adversarial attacks, ordered in increasing complexity: Fast Gradient Sign Method (FGSM) \cite{fgsm}, Projected Gradient Descent (PGD) \cite{pgd}, and Auto-PGD (APGD) \cite{apgd}. We employ PGD and APGD attacks with 100 iterations while FGSM uses single iteration by design. Our evaluation focuses on \(\ell_\infty\) bounded perturbations, with the perturbation magnitudes \(\epsilon \in \{4/255, 8/255, 16/255\}\). This range allows us to systematically assess the robustness of  models against varying strengths of adversarial attacks.

\subsection{Tasks}

Our evaluation covers two primary tasks: Image Captioning and VQA. For image captioning, we use the validation splits of the COCO \cite{COCO} and Flickr30k \cite{flickr} datasets to assess caption accuracy and relevance. In the VQA domain, we evaluate using the validation splits of VQAv2 \cite{vqa}, TextVQA \cite{textvqa}, OK-VQA \cite{okvqa}, and VizWiz \cite{vizwiz} datasets. 
We report the robust VQA accuracy for datasets associated with VQA tasks and robust CIDEr scores for the captioning datasets. Higher is better for both metrics. We randomly sample 1000 examples from the validation set of each task and use this for the adversarial evaluations of all models to ensure a fair comparison. The models selected for evaluating the impact of design choices on adversarial robustness are detailed in Table \ref{tab:models}. 

\begin{table}[!h]
\centering
\caption{Models used for evaluation of various components of VLMs. Each row corresponds to a VLM built with the given vision encoder and LLM. }

\resizebox{\columnwidth}{!}{%
\begin{tabular}{lll}\toprule

\textbf{} & Vision Encoder & Language Model \\ \toprule 
\begin{tabular}[c]{@{}l@{}}Image \\ Representations\end{tabular} & \begin{tabular}[c]{@{}l@{}}CLIP ViT-L/14 @ 224px\\ SigLIP ViT-SO/14 @ 224px\\ DINOv2 ViT-L/14 @ 224px\\ ImageNet-21K+1K ViT-L/16 @ 224px\end{tabular} & Vicuna v1.5 7B \\ \midrule
\begin{tabular}[c]{@{}l@{}}Image \\ Resolution\end{tabular} & \begin{tabular}[c]{@{}l@{}}CLIP ViT-L/14 @ 224px\\ SigLIP ViT-SO/14 @ 224px\\ CLIP ViT-L/14 @ 336px\\ SigLIP ViT-SO/14 @ 384px\end{tabular} & Vicuna v1.5 7B \\ \midrule
Size of LLM & CLIP ViT-L/14 @ 336px & \begin{tabular}[c]{@{}l@{}}Vicuna v1.5 7B\\ Vicuna v1.5 13B\end{tabular} \\ \midrule
\begin{tabular}[c]{@{}l@{}}Ensemble of \\ visual encoders\end{tabular} & \begin{tabular}[c]{@{}l@{}}DINOv2 ViT-L/14 + \\ CLIP ViT-L/14 @ 336px\\ DINOv2 ViT-L/14 + \\ SigLIP ViT-L/14 @ 384px\end{tabular} & Vicuna v1.5 7B \\ \bottomrule
\end{tabular}%
}
\label{tab:models}
\end{table}

\section{Results}


\begin{figure*}
    \vskip 0.2in
    \centering
    \includegraphics[width=\textwidth]{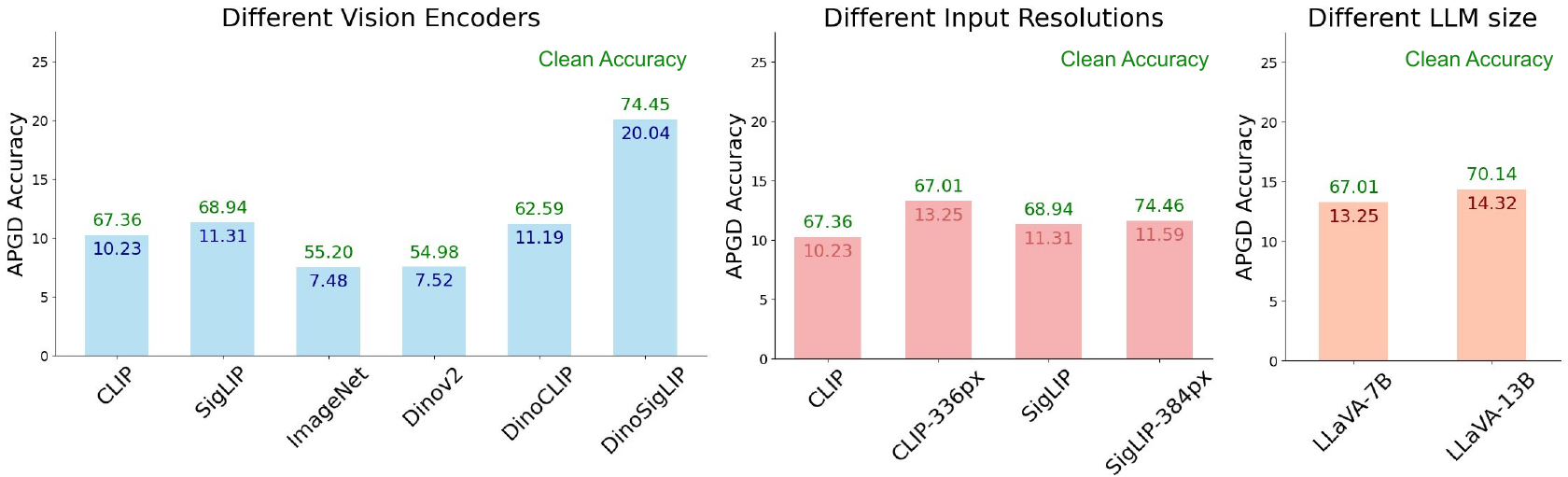}
    \caption{Comparison between VLMs having different vision encoders (left), different input resolutions (center) and different LLM size (right). The comparison is based on the APGD accuracy averaged over all tasks as shown in Tables  \ref{tab:image-encoder-comparison}, \ref{tab:image-encoder-resolution}, \ref{tab:model-scale} and \ref{tab:ensemble-ves}.}
    \label{fig:enter-label}
\end{figure*}

\subsection{Model Design Choices}

In our analysis, we examine the impact of various model design choices on adversarial robustness. 
Specifically, we focus on: (a) the choice of vision encoder; (b) the input resolution used by the vision encoder; (c) the sizes of the language models; and (d) the ensemble use of multiple vision encoders. Each of these aspects is detailed in the sections below. We report results using \(\epsilon = 8/255\). Please check Appendix \ref{appendix:model_design_choice_results} for results with other values \(\epsilon = 4/255, 16/255\). 
We highlight the best robust \colorbox{my_green}{FGSM}, \colorbox{my_blue}{PGD}, \colorbox{my_red}{APGD} accuracy across benchmarks for all attacks and models.


\subsubsection{Impact of Vision Encoder}

We systematically evaluate the effects of employing different vision encoders, each trained under distinct conditions. We compare VLMs that use four different image encoders: CLIP \cite{clip}, SigLIP \cite{siglip}, DINOv2 \cite{dino}, and ImageNet \cite{vit}. As shown in Table \ref{tab:image-encoder-comparison}, SigLIP slightly outperforms CLIP, with both noticeably surpassing DINOv2 and ImageNet VLMs on weaker attacks. However, the difference diminishes for stronger attacks.
We hypothesize that the Vision Transformer (ViT) used in CLIP and SigLIP has been trained across a wide spectrum of internet-collected images and hence has seen many more distributions during training than DINOv2 and ImageNet. 
The results also resonate with the choice of vision encoders in recent state-of-the-art VLMs \cite{llava, prismatic}. 
\setlength{\tabcolsep}{2.5pt}
\begin{table}[!h]
\centering
\caption{Comparison between VLMs having different image encoders but the same language model - Vicuna v1.5 7B. Note: Higher values (\(\uparrow\)) indicate better performance. }

\resizebox{\columnwidth}{!}{%
\begin{tabular}{llrrrrrr} \toprule
\textbf{} & \multicolumn{1}{c}{} & \multicolumn{6}{c}{Task} \\
 & \multicolumn{1}{c}{\multirow{-2}{*}{Attack}} & \multicolumn{1}{c}{COCO} & \multicolumn{1}{c}{Flickr30k} & \multicolumn{1}{c}{OK-VQA} & \multicolumn{1}{c}{TextVQA} & \multicolumn{1}{c}{VizWiz} & \multicolumn{1}{c}{VQav2} \\ \toprule
 & None & 116.35 & 76.15 & 59.48 & 37.10 & 41.18 & 73.89 \\
 & FGSM & \cellcolor[HTML]{E5F3E0}94.18 & \cellcolor[HTML]{E5F3E0}56.83 & 47.58 & 22.40 & 33.70 & 57.85 \\
 & PGD & 13.36 & 9.11 & 13.90 & 7.42 & 8.67 & 31.65 \\
\multirow{-4}{*}{CLIP} & APGD & 6.32 & 4.41 & 10.11 & 4.80 & 8.16 & 27.56 \\ \midrule
 & None & 118.69 & 78.29 & 60.92 & 38.82 & 42.43 & 74.49 \\
 & FGSM & 94.09 & 54.43 & \cellcolor[HTML]{E5F3E0}48.24 & \cellcolor[HTML]{E5F3E0}24.38 & \cellcolor[HTML]{E5F3E0}39.85 & \cellcolor[HTML]{E5F3E0}60.63 \\
 & PGD & \cellcolor[HTML]{DBECFC}20.46 & \cellcolor[HTML]{DBECFC}11.03 & \cellcolor[HTML]{DBECFC}14.96 & \cellcolor[HTML]{DBECFC}7.52 & \cellcolor[HTML]{DBECFC}10.91 & \cellcolor[HTML]{DBECFC}34.70 \\
\multirow{-4}{*}{SigLIP} & APGD & \cellcolor[HTML]{F0DAE3}7.32 & \cellcolor[HTML]{F0DAE3}4.87 & \cellcolor[HTML]{F0DAE3}10.44 & \cellcolor[HTML]{F0DAE3}5.30 & 9.54 & \cellcolor[HTML]{F0DAE3}30.38 \\ \midrule
 & None & 104.84 & 54.78 & 57.00 & 10.37 & 38.07 & 64.80 \\
 & FGSM & 81.81 & 38.24 & 40.08 & 8.03 & 33.78 & 46.24 \\
 & PGD & 3.07 & 1.80 & 9.06 & 1.83 & 10.60 & 25.21 \\
\multirow{-4}{*}{DINOv2} & APGD & 2.09 & 1.22 & 7.16 & 2.00 & 10.20 & 22.47 \\ \midrule
 & None & 101.59 & 54.92 & 56.34 & 10.70 & 39.29 & 68.36 \\
 & FGSM & 69.13 & 32.42 & 38.40 & 6.60 & 32.45 & 46.13 \\
 & PGD & 5.38 & 3.43 & 9.26 & 1.79 & 10.78 & 22.73 \\
\multirow{-4}{*}{In1k} & APGD & 2.74 & 2.04 & 7.58 & 1.52 & \cellcolor[HTML]{F0DAE3}10.22 & 20.79 \\ \bottomrule
\end{tabular}%
}

\label{tab:image-encoder-comparison}
\end{table}

\subsubsection{Resolution of Vision Encoder}
Generally, a higher input resolution improves the quality of visual representations, potentially boosting model performance \cite{prismatic}. Owing to the availability of high-resolution variants, we specifically evaluate models equipped with CLIP and SigLIP vision encoders at two distinct resolutions to thoroughly understand these effects. Based on Table \ref{tab:image-encoder-resolution}, while increasing resolution enhances robustness against stronger attacks for high-resolution CLIP models (on most tasks), the effectiveness of the increased resolution in SigLIP models appears to be task-dependent. 
However, we observe that robust accuracy significantly deteriorates under APGD attacks in all cases except for VQAv2. For results on other \(\epsilon\) values, please refer to Appendix \ref{appendix:model_design_choice_results}.

\begin{table}[!h]
\centering
\caption{Comparison between VLMs having different input resolutions of CLIP and SigLIP. All of them have the same language model: Vicuna v1.5 7B. Note: Higher values (\(\uparrow\)) indicate better performance. }

\resizebox{\columnwidth}{!}{%
\begin{tabular}{llrrrrrr} \toprule
\textbf{} & \multicolumn{1}{c}{} & \multicolumn{6}{c}{Task} \\
 & \multicolumn{1}{c}{\multirow{-2}{*}{Attack}} & \multicolumn{1}{l}{COCO} & \multicolumn{1}{l}{Flickr30k} & \multicolumn{1}{l}{OK-VQA} & \multicolumn{1}{l}{TextVQA} & \multicolumn{1}{l}{VizWiz} & \multicolumn{1}{l}{VQav2} \\ \toprule
 & None & 116.35 & 76.15 & 59.48 & 37.10 & 41.18 & 73.89 \\
 & FGSM & 94.18 & 56.83 & 47.58 & 22.40 & 33.70 & \cellcolor[HTML]{E5F3E0}57.85 \\
 & PGD & 13.36 & 9.11 & 13.90 & 7.42 & 8.67 & \cellcolor[HTML]{DBECFC}31.65 \\
\multirow{-4}{*}{CLIP-224px} & APGD & 6.32 & 4.41 & 10.11 & 4.80 & 8.16 & \cellcolor[HTML]{F0DAE3}27.56 \\ \midrule
 & None & 119.02 & 77.21 & 59.18 & 36.73 & 39.94 & 70.00 \\
 & FGSM & \cellcolor[HTML]{E5F3E0}95.55 & \cellcolor[HTML]{E5F3E0}63.10 & \cellcolor[HTML]{E5F3E0}48.20 & \cellcolor[HTML]{E5F3E0}24.03 & \cellcolor[HTML]{E5F3E0}35.24 & 57.19 \\
 & PGD & \cellcolor[HTML]{DBECFC}22.84 & \cellcolor[HTML]{DBECFC}13.87 & \cellcolor[HTML]{DBECFC}20.46 & \cellcolor[HTML]{DBECFC}10.13 & \cellcolor[HTML]{DBECFC}22.73 & 27.07 \\
\multirow{-4}{*}{CLIP-336px} & APGD & \cellcolor[HTML]{F0DAE3}12.54 & \cellcolor[HTML]{F0DAE3}7.15 & \cellcolor[HTML]{F0DAE3}15.68 & \cellcolor[HTML]{F0DAE3}8.08 & \cellcolor[HTML]{F0DAE3}9.31 & 26.73 \\ \midrule
 \midrule
 & None & 118.69 & 78.29 & 60.92 & 38.82 & 42.43 & 74.49 \\
 & FGSM & \cellcolor[HTML]{E5F3E0}94.09 & 54.43 & 48.24 & 24.38 & \cellcolor[HTML]{E5F3E0}39.85 & 60.63 \\
 & PGD & \cellcolor[HTML]{DBECFC}20.46 & \cellcolor[HTML]{DBECFC}11.03 & 14.96 & 7.52 & \cellcolor[HTML]{DBECFC}10.91 & 34.70 \\
\multirow{-4}{*}{SigLIP-224px} & APGD & \cellcolor[HTML]{F0DAE3}7.32 & \cellcolor[HTML]{F0DAE3}4.87 & 10.44 & 5.30 & \cellcolor[HTML]{F0DAE3}9.54 & 30.38 \\ \midrule
 & None & 124.11 & 87.08 & 62.18 & 55.05 & 41.14 & 77.22 \\
 & FGSM & 92.39 & \cellcolor[HTML]{E5F3E0}57.55 & \cellcolor[HTML]{E5F3E0}51.38 & \cellcolor[HTML]{E5F3E0}32.91 & 37.28 & \cellcolor[HTML]{E5F3E0}62.47 \\
 & PGD & 15.69 & 8.29 & \cellcolor[HTML]{DBECFC}18.04 & \cellcolor[HTML]{DBECFC}9.61 & 9.98 & \cellcolor[HTML]{DBECFC}35.97 \\
\multirow{-4}{*}{SigLIP-384px} & APGD & 6.90 & 3.22 & \cellcolor[HTML]{F0DAE3}12.72 & \cellcolor[HTML]{F0DAE3}6.73 & 8.77 & \cellcolor[HTML]{F0DAE3}31.21 \\ \bottomrule
\end{tabular}%
}
\label{tab:image-encoder-resolution}
\end{table}

\subsubsection{Size of language model}
We evaluate a series of VLMs utilizing the same vision encoder, and same LLM architecture, with the only difference being the language model's size.
Specifically, we examine models equipped with the Vicuna language model \cite{vicuna} in two sizes: 7B and 13B. 
According to the results in Table \ref{tab:model-scale}, the model's vulnerability to adversarial attacks and the significant drop in robust accuracy remain consistent, regardless of the model’s scale.
Hence, increasing the size of the language model does not seem to enhance robustness. One potential reason for this could be that adversarial attacks compromise the representations from the vision encoder.
As a result, LLMs even at 13B scale may struggle to effectively interpret these flawed representations, making robustness to image-based attacks less sensitive to language model size.
Therefore, enhancing the vision encoder's adversarial robustness is sufficient as shown in prior work \cite{robustclip}. Please check Appendix \ref{appendix:model_design_choice_results} for results on other \(\epsilon\) values.\looseness-1
\begin{table}[!h]
\centering
\caption{Comparison between models having different scales of language models. Both of the models have the same vision encoder, CLIP-336, but different scales of the LLM. Note: Higher values (\(\uparrow\)) indicate better performance.}

\resizebox{\columnwidth}{!}{%
\begin{tabular}{llrrrrrr} \toprule
\textbf{} & \multicolumn{1}{c}{} & \multicolumn{6}{c}{Task} \\
 & \multicolumn{1}{c}{\multirow{-2}{*}{Attack}} & \multicolumn{1}{c}{COCO} & \multicolumn{1}{c}{Flickr30k} & \multicolumn{1}{c}{OK-VQA} & \multicolumn{1}{c}{TextVQA} & \multicolumn{1}{c}{VizWiz} & \multicolumn{1}{c}{VQav2} \\ \toprule
 & None & 119.02 & 77.21 & 59.18 & 36.73 & 39.94 & 70.00 \\
 & FGSM & 95.55 & 63.10 & 48.20 & 24.03 & 35.24 & 57.19 \\
 & PGD & \cellcolor[HTML]{DBECFC}22.84 & \cellcolor[HTML]{DBECFC}13.87 & \cellcolor[HTML]{DBECFC}20.46 & \cellcolor[HTML]{DBECFC}10.13 & 22.73 & 27.07 \\
\multirow{-4}{*}{LLaVA-7B} & APGD & \cellcolor[HTML]{F0DAE3}12.54 & \cellcolor[HTML]{F0DAE3}7.15 & 15.68 & 8.08 & \cellcolor[HTML]{F0DAE3}9.31 & 26.73 \\ \midrule
 & None & 123.71 & 77.63 & 62.86 & 40.04 & 41.19 & 75.39 \\
 & FGSM & \cellcolor[HTML]{E5F3E0}106.40 & \cellcolor[HTML]{E5F3E0}64.93 & \cellcolor[HTML]{E5F3E0}50.90 & \cellcolor[HTML]{E5F3E0}26.28 & \cellcolor[HTML]{E5F3E0}36.48 & \cellcolor[HTML]{E5F3E0}62.49 \\
 & PGD & 14.60 & 8.96 & 15.65 & 9.08 & \cellcolor[HTML]{DBECFC}23.55 & \cellcolor[HTML]{DBECFC}34.49 \\
\multirow{-4}{*}{LLaVA-13B} & APGD & 6.98 & 4.35 & \cellcolor[HTML]{F0DAE3}23.13 & \cellcolor[HTML]{F0DAE3}10.45 & 7.47 & \cellcolor[HTML]{F0DAE3}33.56 \\ \bottomrule
\end{tabular}%
}
\label{tab:model-scale}
\end{table}

\subsubsection{Ensemble of vision encoders}
We also explore the vulnerability of VLMs that employ an ensemble of vision encoders. Although recent studies suggest that multiple encoders can significantly improve performance \cite{prismatic, brave}, our research aims to assess whether compromising just one encoder can affect the entire model.
This approach allows us to analyze if knowledge of the weakest link is sufficient to compromise the entire model when an ensemble of encoders is used. We specifically examined models combining DINOv2 with either CLIP or SigLIP. In our experiments, we perturbed the images processed by DINOv2 while keeping inputs to the other encoder intact.  Results in Table \ref{tab:ensemble-ves}
 show that attacking only DINOv2 is sufficient to compromise the model under stronger attacks, despite the other vision encoder providing clean inputs. This highlights a significant vulnerability in ensemble approaches: even with enhanced performance capabilities, the robustness of the entire system can be jeopardized by targeting a single encoder. Please check Appendix \ref{appendix:model_design_choice_results} for results on other \(\epsilon\) values.

\begin{table}[!h]
\centering
\caption{Comparison between VLMs that have an ensemble of vision encoders. The comparison is made when only the input to the Dino image encoder is perturbed. Note: Higher values (\(\uparrow\)) indicate better performance. }

\resizebox{\columnwidth}{!}{%
\begin{tabular}{llrrrrrr} \toprule
\textbf{} & \multicolumn{1}{c}{} & \multicolumn{6}{c}{Task} \\
 & \multicolumn{1}{c}{\multirow{-2}{*}{Attack}} & \multicolumn{1}{c}{COCO} & \multicolumn{1}{c}{Flickr30k} & \multicolumn{1}{c}{OK-VQA} & \multicolumn{1}{c}{TextVQA} & \multicolumn{1}{c}{VizWiz} & \multicolumn{1}{c}{VQav2} \\ \toprule
 & None & 113.75 & 74.16 & 58.88 & 15.08 & 39.30 & 74.35 \\
 & FGSM & \cellcolor[HTML]{E5F3E0}100.75 & 55.29 & 47.54 & 8.98 & 39.11 & 58.20 \\
 & PGD & 13.14 & 7.40 & 12.44 & 3.11 & 12.94 & 33.32 \\
\multirow{-4}{*}{Dino+CLIP} & APGD & 5.99 & 4.36 & 11.08 & 2.88 & 12.51 & 30.29 \\ \midrule
 & None & 125.94 & 85.44 & 61.12 & 50.52 & 44.27 & 79.39 \\
 & FGSM & 109.79 & \cellcolor[HTML]{E5F3E0}73.84 & \cellcolor[HTML]{E5F3E0}53.94 & \cellcolor[HTML]{E5F3E0}40.84 & \cellcolor[HTML]{E5F3E0}41.84 & \cellcolor[HTML]{E5F3E0}67.75 \\
 & PGD & \cellcolor[HTML]{DBECFC}39.76 & \cellcolor[HTML]{DBECFC}22.64 & \cellcolor[HTML]{DBECFC}19.30 & \cellcolor[HTML]{DBECFC}12.83 & \cellcolor[HTML]{DBECFC}13.37 & \cellcolor[HTML]{DBECFC}39.78 \\
\multirow{-4}{*}{Dino+SigLIP} & APGD & \cellcolor[HTML]{F0DAE3}25.23 & \cellcolor[HTML]{F0DAE3}15.72 & \cellcolor[HTML]{F0DAE3}17.26 & \cellcolor[HTML]{F0DAE3}12.03 & \cellcolor[HTML]{F0DAE3}12.25 & \cellcolor[HTML]{F0DAE3}37.75 \\ 
\bottomrule
\end{tabular}%
}
\label{tab:ensemble-ves}
\end{table}

\subsection{Prompt Formatting}

Considering that our adversarial examples are generated solely by perturbing visual inputs, we hypothesize that modifying the original prompts could be particularly effective in countering the effects of such perturbations.
We test this hypothesis with the LLaVA 7B and LLaVA 13B models, employing different types of prompts for COCO and VQAv2. Our evaluation includes adversarial examples created using FGSM, PGD, and APGD attacks, with PGD, APGD based on 100 iterations. Specific details on the prompts used and the corresponding results are detailed in the subsequent subsections.

\subsubsection{Captioning}

Our experiments evaluated various prompt formatting strategies, including: (1) \textbf{Original} - using the original prompt; (2) \textbf{Adversarial Certainty (AC) Prompt} - explicitly informing the model that the image is adversarially perturbed; (3) \textbf{Adversarial Possibility (AP) Prompt} - suggesting the possibility that the image might be adversarially perturbed; and (4) \textbf{Random} - appending a random sentence or string at the beginning of the prompt. These are listed in Table \ref{tab:prompts-captioning} in Appendix \ref{appendix:prompt_formatting}. From the results presented in Fig. \ref{fig:prompt_captioning} and Table \ref{tab:captioning-prompting} in Appendix \ref{appendix:prompt_formatting_results}, it is evident that indicating the possibility of adversarial perturbations (AP prompt) assists the model significantly more than explicitly stating that the image is perturbed (AC prompt). Further, the improvements from simply adding a random string or sentence are substantial, even comparable to the effects observed with the AP prompt. This indicates that the models pay more attention to the inputs when they struggle to establish a clear relationship between them.\looseness-1



\subsubsection{Visual Question Answering}
Here, we explored four strategies: (1) \textbf{Rephrase} - rephrasing the original question to create a semantically similar question; (2) \textbf{Expand} - increasing the length of the questions; (3) \textbf{Adversarial Certainty (AC) Prompt} - explicitly informing the model that the image is adversarially perturbed; and (4) \textbf{Adversarial Possibility (AP) Prompt} - suggesting the possibility that the image might be adversarially perturbed. We utilize a finetuned model of the Mistral 7B LLM \cite{mistral} to generate questions according to the above-mentioned strategies.
All the instructions used to obtain the modified questions are listed in Table \ref{tab:vqa-prompts} in Appendix \ref{appendix:prompt_formatting}. According to the results presented in Fig. \ref{fig:vqa_prompt_fig} and Table \ref{tab:vqa-prompting} in Appendix \ref{appendix:prompt_formatting_results}, simply rephrasing the questions significantly improved performance compared to the other methods, such as extending the question length or explicitly warning about potential adversarial perturbations. Moreover, indicating the possibility of an adversarial perturbation yielded the best robustness performance, reinforcing our observations with the COCO dataset discussed earlier.

\begin{figure}
    \vskip 0.2in
    \centering
    \includegraphics[width=\columnwidth]{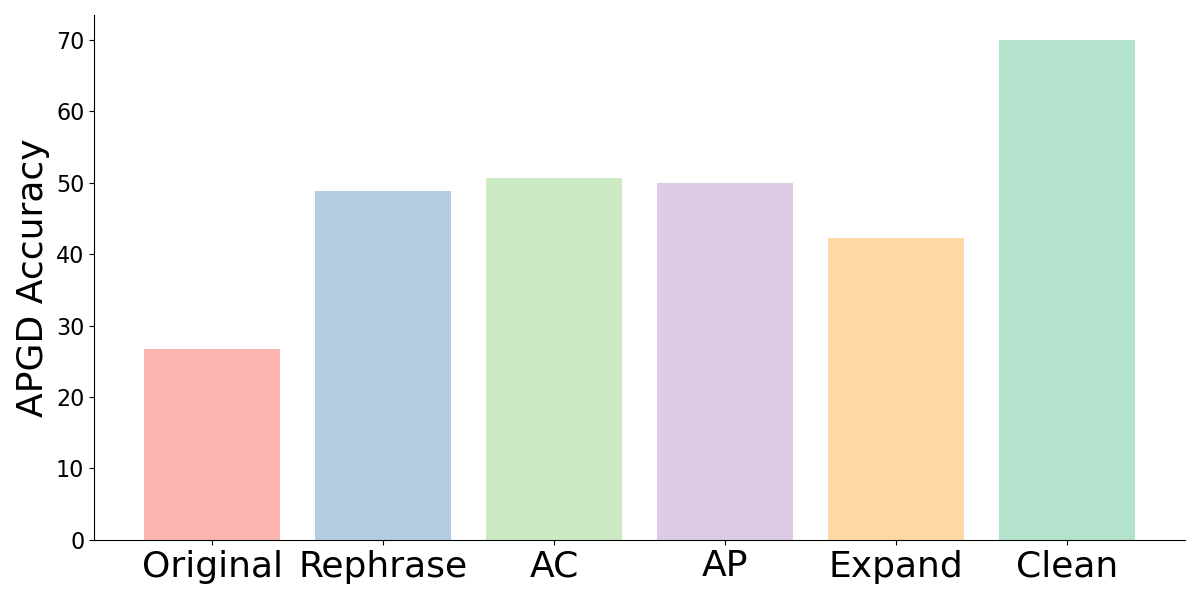}
    \caption{Performance of LLaVA-7B on VQA using questions generated by different types of prompts.}
    \label{fig:vqa_prompt_fig}
\end{figure}

\section{Discussion and Conclusion}


Our evaluation highlighted critical insights into how various design elements affect the adversarial robustness of VLMs. First, we observed that vision encoders trained across diverse data distributions only improve resistance against simpler, less sophisticated attacks, demonstrating limited effectiveness against more complex threats. Additionally, increasing the resolution of image encoders did not correlate with enhanced adversarial robustness, suggesting that benefits seen in the clean accuracy do not extend to improved robustness. Similarly, scaling up the size of the language model did not increase the model’s robustness to attacks, indicating that larger models are not inherently more robust. Most notably, our results revealed that using multiple vision encoders does not guarantee robustness; rather, knowledge about the most vulnerable encoder is enough to compromise the entire system.

Building on our findings, we further explored the influence of prompt formatting on enhancing adversarial robustness. Our experiments revealed that even naively rephrasing the questions significantly improves robustness in VQA. Similarly, merely suggesting the possibility of an adversarial image during captioning led to a notable performance boost. More importantly, we found that we do not need to add additional context from the image or fine-tune additional tokens to make models adversarially robust, as opposed to prior work \cite{useful_context, adv-prompt}. 

These findings underscore the critical impact of model design and prompt formulation on a model's robustness to adversarial attacks, demonstrating that even minimal modifications to the textual prompt can significantly enhance the model's robustness against visual attacks.

\section{Impact Statement}
As VLMs see increased real-world deployment, ensuring their robustness against adversarial attacks is critical. Our research makes two key contributions: providing optimal model design choices for safe deployment and demonstrating how prompt formatting can enhance adversarial robustness. Our lightweight technique offers a practical alternative to computationally intensive adversarial training, reducing the computational footprint. While enhancing robustness against multimodal attacks using prompt formatting remains unexplored, our work addresses the crucial task of defending against strong image-based attacks that can lead to misinformation or harmful content generation. This research aims to support future advancements in the safe deployment of AI systems.

\section{Acknowledgements}

We acknowledge support from the Canada CIFAR AI Chair Program and from the Canada Excellence Research Chairs Program. P.B. was supported by FRQ-S Research Scholars Junior 1 grant 310924, and the William Dawson Scholar award. This research was enabled in part by computational resources provided by the Digital Research Alliance of Canada and Mila Quebec AI Institute. 





\bibliography{example_paper}
\bibliographystyle{icml2024}

\newpage
\appendix
\onecolumn
\section{Model Design Choice Results}
\label{appendix:model_design_choice_results}
We provide results for studying the impact of various model design choices for \(\epsilon = 4/255\) and \(16/255\) here. 

\subsection{Impact of Vision Encoder}
We can observe that for a lower \(\epsilon\) value, i.e., \(4/255\) CLIP performs better. However, for higher \(\epsilon\) values, i.e. \(8/255\) and \(16/255\), SigLIP performs better. 

\begin{table}[!h]
\setlength{\tabcolsep}{0.51cm}
\centering
\caption{Comparison between VLMs having different image encoders but the same language model for \(\epsilon = 4/255\). All of them have the same language model: Vicuna v1.5 7B. Note: Higher values (\(\uparrow\)) indicate better performance.}
\vspace{-2mm}
\resizebox{\columnwidth}{!}{%
\begin{tabular}{llrrrrrr}\toprule
 &  & \multicolumn{6}{c}{Task} \\
 & \multirow{-2}{*}{Attack} & \multicolumn{1}{c}{COCO} & \multicolumn{1}{c}{Flickr30k} & \multicolumn{1}{c}{OK-VQA} & \multicolumn{1}{c}{TextVQA} & \multicolumn{1}{c}{VizWiz} & \multicolumn{1}{c}{VQAv2} \\ \toprule
 & None & 116.35 & 76.15 & 59.48 & 37.10 & 41.18 & 73.89 \\
 & FGSM & \cellcolor[HTML]{E5F3E0}106.01 & \cellcolor[HTML]{E5F3E0}64.45 & \cellcolor[HTML]{E5F3E0}52.18 & 26.87 & 36.93 & \cellcolor[HTML]{E5F3E0}63.55 \\
 & PGD & \cellcolor[HTML]{DBECFC}89.95 & \cellcolor[HTML]{DBECFC}54.54 & \cellcolor[HTML]{DBECFC}44.40 & 6.73 & \cellcolor[HTML]{DBECFC}32.06 & \cellcolor[HTML]{DBECFC}53.81 \\
\multirow{-4}{*}{CLIP} & APGD & \cellcolor[HTML]{F0DAE3}87.07 & \cellcolor[HTML]{F0DAE3}50.51 & \cellcolor[HTML]{F0DAE3}42.52 & \cellcolor[HTML]{F0DAE3}19.03 & 8.80 & \cellcolor[HTML]{F0DAE3}50.16 \\ \midrule
 & None & 118.69 & 78.29 & 60.92 & 38.82 & 42.43 & 74.49 \\
 & FGSM & 99.75 & 60.60 & 49.84 & \cellcolor[HTML]{E5F3E0}27.24 & \cellcolor[HTML]{E5F3E0}39.74 & 61.75 \\
 & PGD & 68.94 & 38.42 & 33.30 & \cellcolor[HTML]{DBECFC}9.54 & 26.65 & 44.51 \\
\multirow{-4}{*}{SigLIP} & APGD & 59.67 & 33.12 & 14.45 & 11.89 & \cellcolor[HTML]{F0DAE3}24.12 & 41.87 \\ \midrule
 & None & 104.84 & 54.78 & 57.00 & 10.37 & 38.07 & 64.80 \\
 & FGSM & 77.68 & 37.81 & 39.78 & 7.28 & 32.50 & 45.40 \\
 & PGD & 4.86 & 3.13 & 9.80 & 1.99 & 10.91 & 25.67 \\
\multirow{-4}{*}{Dinov2} & APGD & 2.45 & 2.17 & 8.00 & 1.96 & 10.69 & 23.29 \\ \midrule
 & None & 101.59 & 54.92 & 56.34 & 10.70 & 39.29 & 68.36 \\
 & FGSM & 71.67 & 34.74 & 38.44 & 6.70 & 31.62 & 45.37 \\
 & PGD & 11.17 & 5.62 & 11.28 & 2.43 & 11.90 & 15.00 \\
\multirow{-4}{*}{ImageNet} & APGD & 5.24 & 3.69 & 9.86 & 2.04 & 10.80 & 17.14 \\ \bottomrule
\end{tabular}%
}
\label{tab:comparison_ve_eps_4}
\end{table}

\begin{table}[!h]
\setlength{\tabcolsep}{0.51cm}
\centering
\caption{Comparison between VLMs having different image encoders but the same language model for \(\epsilon = 16/255\). All of them have the same language model: Vicuna v1.5 7B. Note: Higher values (\(\uparrow\)) indicate better performance.}
\vspace{-2mm}
\resizebox{\columnwidth}{!}{%
\begin{tabular}{llrrrrrr} \toprule
 &  & \multicolumn{6}{c}{Task} \\
 & \multirow{-2}{*}{Attack} & \multicolumn{1}{c}{COCO} & \multicolumn{1}{c}{Flickr30k} & \multicolumn{1}{c}{OK-VQA} & \multicolumn{1}{c}{TextVQA} & \multicolumn{1}{c}{VizWiz} & \multicolumn{1}{c}{VQAv2} \\ \toprule
 & None & 116.35 & 76.15 & 59.48 & 37.10 & 41.18 & 73.89 \\
 & FGSM & \cellcolor[HTML]{E5F3E0}93.27 & \cellcolor[HTML]{E5F3E0}56.35 & 48.52 & 20.22 & 36.73 & 59.99 \\
 & PGD & 10.32 & 6.22 & 11.88 & 5.87 & 8.23 & 29.57 \\
\multirow{-4}{*}{CLIP} & APGD & 3.33 & 2.57 & \cellcolor[HTML]{F0DAE3}8.40 & \cellcolor[HTML]{F0DAE3}3.84 & 7.89 & \cellcolor[HTML]{F0DAE3}23.84 \\ \midrule
 & None & 118.69 & 78.29 & 60.92 & 38.82 & 42.43 & 74.49 \\
 & FGSM & 88.06 & 50.05 & \cellcolor[HTML]{E5F3E0}48.62 & \cellcolor[HTML]{E5F3E0}22.22 & \cellcolor[HTML]{E5F3E0}40.80 & \cellcolor[HTML]{E5F3E0}60.55 \\
 & PGD & \cellcolor[HTML]{DBECFC}11.77 & \cellcolor[HTML]{DBECFC}6.59 & \cellcolor[HTML]{DBECFC}12.45 & \cellcolor[HTML]{DBECFC}6.57 & \cellcolor[HTML]{DBECFC}10.32 & \cellcolor[HTML]{DBECFC}31.04 \\
\multirow{-4}{*}{SigLIP} & APGD & \cellcolor[HTML]{F0DAE3}4.31 & \cellcolor[HTML]{F0DAE3}2.79 & 7.88 & 3.78 & 8.97 & 23.50 \\ \midrule
 & None & 104.84 & 54.78 & 57.00 & 10.37 & 38.07 & 64.80 \\
 & FGSM & 81.38 & 39.35 & 41.18 & 7.96 & 36.22 & 48.61 \\
 & PGD & 3.00 & 1.48 & 7.70 & 1.54 & 10.68 & 24.78 \\
\multirow{-4}{*}{Dinov2} & APGD & 1.57 & 1.12 & 6.34 & 1.34 & 9.70 & 20.73 \\ \midrule
 & None & 101.59 & 54.92 & 56.34 & 10.70 & 39.29 & 68.36 \\
 & FGSM & 62.62 & 29.90 & 39.42 & 7.20 & 33.98 & 46.78 \\
 & PGD & 3.12 & 2.13 & 8.10 & 1.64 & 9.34 & 22.69 \\
\multirow{-4}{*}{ImageNet} & APGD & 2.13 & 0.95 & 5.84 & 1.80 & \cellcolor[HTML]{F0DAE3}9.98 & 18.99 \\ \bottomrule
\end{tabular}
}
\label{tab:comparison_ve_eps_16}
\end{table}

\subsection{Resolution of Vision Encoder}

We can observe that at a lower \(\epsilon\) value of \(4/255\), lower resolution models are better. However, at a higher \(\epsilon\) value of \(16/255\), the effectiveness of increased resolution for both CLIP and SigLIP models becomes task-dependent.

\begin{table}[!h]
\setlength{\tabcolsep}{0.5cm}
\centering
\caption{Comparison between VLMs having different input resolutions of CLIP and SigLIP for \(\epsilon = 4/255\). All of them have the same language model: Vicuna v1.5 7B. Note: Higher values (\(\uparrow\)) indicate better performance.}
\vspace{-2mm}
\resizebox{\columnwidth}{!}{%
\begin{tabular}{llrrrrrr} \toprule
 &  & \multicolumn{6}{c}{Task} \\
 & \multirow{-2}{*}{Attack} & \multicolumn{1}{c}{COCO} & \multicolumn{1}{c}{Flickr30k} & \multicolumn{1}{c}{OK-VQA} & \multicolumn{1}{c}{TextVQA} & \multicolumn{1}{c}{VizWiz} & \multicolumn{1}{c}{VQAv2} \\ \toprule
 & None & 116.35 & 76.15 & 59.48 & 37.10 & 41.18 & 73.89 \\
 & FGSM & \cellcolor[HTML]{E5F3E0}106.01 & \cellcolor[HTML]{E5F3E0}64.45 & \cellcolor[HTML]{E5F3E0}52.18 & \cellcolor[HTML]{E5F3E0}26.87 & \cellcolor[HTML]{E5F3E0}36.93 & \cellcolor[HTML]{E5F3E0}63.55 \\
 & PGD & \cellcolor[HTML]{DBECFC}89.95 & \cellcolor[HTML]{DBECFC}54.54 & \cellcolor[HTML]{DBECFC}44.40 & 6.73 & \cellcolor[HTML]{DBECFC}32.06 & \cellcolor[HTML]{DBECFC}53.81 \\
\multirow{-4}{*}{CLIP-224px} & APGD & \cellcolor[HTML]{F0DAE3}87.07 & \cellcolor[HTML]{F0DAE3}50.51 & \cellcolor[HTML]{F0DAE3}42.52 & \cellcolor[HTML]{F0DAE3}19.03 & 8.80 & \cellcolor[HTML]{F0DAE3}50.16 \\ \midrule
 & None & 119.02 & 77.21 & 59.18 & 36.73 & 39.94 & 70.00 \\
 & FGSM & 96.79 & 64.25 & 49.70 & 25.32 & 33.92 & 56.52 \\
 & PGD & 18.54 & 13.57 & 16.26 & \cellcolor[HTML]{DBECFC}9.44 & 10.67 & 28.60 \\
\multirow{-4}{*}{CLIP-336px} & APGD & 30.20 & 22.11 & 22.86 & 11.73 & \cellcolor[HTML]{F0DAE3}22.86 & 28.76 \\ \midrule \midrule
 & None & 118.69 & 78.29 & 60.92 & 38.82 & 42.43 & 74.49 \\
 & FGSM & \cellcolor[HTML]{E5F3E0}99.75 & \cellcolor[HTML]{E5F3E0}60.60 & 49.84 & 27.24 & \cellcolor[HTML]{E5F3E0}39.74 & 61.75 \\
 & PGD & \cellcolor[HTML]{DBECFC}68.94 & \cellcolor[HTML]{DBECFC}38.42 & \cellcolor[HTML]{DBECFC}33.30 & 9.54 & \cellcolor[HTML]{DBECFC}26.65 & \cellcolor[HTML]{DBECFC}44.51 \\
\multirow{-4}{*}{SigLIP-224px} & APGD & \cellcolor[HTML]{F0DAE3}59.67 & \cellcolor[HTML]{F0DAE3}33.12 & 14.45 & \cellcolor[HTML]{F0DAE3}11.89 & \cellcolor[HTML]{F0DAE3}24.12 & \cellcolor[HTML]{F0DAE3}41.87 \\ \midrule
 & None & 124.11 & 87.08 & 62.18 & 55.05 & 41.14 & 77.22 \\
 & FGSM & 93.46 & 57.28 & \cellcolor[HTML]{E5F3E0}50.88 & \cellcolor[HTML]{E5F3E0}36.18 & 35.50 & \cellcolor[HTML]{E5F3E0}63.22 \\
 & PGD & 25.51 & 13.84 & 20.32 & \cellcolor[HTML]{DBECFC}13.38 & 11.51 & 38.15 \\
\multirow{-4}{*}{SigLIP-384px} & APGD & 12.53 & 8.40 & \cellcolor[HTML]{F0DAE3}16.34 & 9.10 & 9.52 & 34.83 \\ \bottomrule
\end{tabular}%
}
\label{tab:comparison_ip_res_eps_4}
\end{table}

\begin{table}[!h]
\setlength{\tabcolsep}{0.5cm}
\centering
\caption{Comparison between VLMs having different input resolutions of CLIP and SigLIP for  \(\epsilon = 16/255\). All of them have the same language model: Vicuna v1.5 7B. Note: Higher values (\(\uparrow\)) indicate better performance.}
\vspace{-2mm}
\resizebox{\columnwidth}{!}{%
\begin{tabular}{llrrrrrr} \toprule
 &  & \multicolumn{6}{c}{Task} \\
 & \multirow{-2}{*}{Attack} & \multicolumn{1}{c}{COCO} & \multicolumn{1}{c}{Flickr30k} & \multicolumn{1}{c}{OK-VQA} & \multicolumn{1}{c}{TextVQA} & \multicolumn{1}{c}{VizWiz} & \multicolumn{1}{c}{VQAv2} \\ \toprule
 & None & 116.35 & 76.15 & 59.48 & 37.10 & 41.18 & 73.89 \\
 & FGSM & 93.27 & \cellcolor[HTML]{E5F3E0}56.35 & 48.52 & 20.22 & 36.73 & \cellcolor[HTML]{E5F3E0}59.99 \\
 & PGD & \cellcolor[HTML]{DBECFC}10.32 & \cellcolor[HTML]{DBECFC}6.22 & 11.88 & 5.87 & 8.23 & \cellcolor[HTML]{DBECFC}29.57 \\
\multirow{-4}{*}{CLIP-224px} & APGD & 3.33 & 2.57 & 8.40 & 3.84 & 7.89 & \cellcolor[HTML]{F0DAE3}23.84 \\ \midrule
 & None & 119.02 & 77.21 & 59.18 & 36.73 & 39.94 & 70.00 \\
 & FGSM & \cellcolor[HTML]{E5F3E0}101.98 & 56.08 & \cellcolor[HTML]{E5F3E0}49.24 & \cellcolor[HTML]{E5F3E0}22.20 & \cellcolor[HTML]{E5F3E0}37.39 & 58.30 \\
 & PGD & 10.08 & 5.25 & \cellcolor[HTML]{DBECFC}17.48 & \cellcolor[HTML]{DBECFC}6.70 & \cellcolor[HTML]{DBECFC}8.52 & 24.39 \\
\multirow{-4}{*}{CLIP-336px} & APGD & \cellcolor[HTML]{F0DAE3}13.88 & \cellcolor[HTML]{F0DAE3}9.05 & \cellcolor[HTML]{F0DAE3}13.26 & \cellcolor[HTML]{F0DAE3}7.34 & \cellcolor[HTML]{F0DAE3}27.54 & 19.98 \\
 \midrule \midrule \\
 & None & 118.69 & 78.29 & 60.92 & 38.82 & 42.43 & 74.49 \\
 & FGSM & 88.06 & 50.05 & 48.62 & 22.22 & \cellcolor[HTML]{E5F3E0}40.80 & 60.55 \\
 & PGD & \cellcolor[HTML]{DBECFC}11.77 & \cellcolor[HTML]{DBECFC}6.59 & 12.45 & 6.57 & \cellcolor[HTML]{DBECFC}10.32 & 31.04 \\
\multirow{-4}{*}{SigLIP-224px} & APGD & \cellcolor[HTML]{F0DAE3}4.31 & \cellcolor[HTML]{F0DAE3}2.79 & 7.88 & 3.78 & \cellcolor[HTML]{F0DAE3}8.97 & 23.50 \\ \midrule
 & None & 124.11 & 87.08 & 62.18 & 55.05 & 41.14 & 77.22 \\
 & FGSM & \cellcolor[HTML]{E5F3E0}94.90 & \cellcolor[HTML]{E5F3E0}57.40 & \cellcolor[HTML]{E5F3E0}52.24 & \cellcolor[HTML]{E5F3E0}30.93 & 39.77 & \cellcolor[HTML]{E5F3E0}63.53 \\
 & PGD & 9.53 & 5.17 & \cellcolor[HTML]{DBECFC}14.48 & \cellcolor[HTML]{DBECFC}8.26 & 9.19 & \cellcolor[HTML]{DBECFC}33.16 \\
\multirow{-4}{*}{SigLIP-384px} & APGD & 3.15 & 1.75 & \cellcolor[HTML]{F0DAE3}9.14 & \cellcolor[HTML]{F0DAE3}4.00 & 7.85 & \cellcolor[HTML]{F0DAE3}27.82 \\ \bottomrule
\end{tabular}%
}
\label{tab:comparison_ip_res_eps_16}
\end{table}

\newpage
\subsection{Size of Language Model}

Here we can observe that increasing the model size only helps in gaining robustness against weaker attacks (FGSM). However, the vulnerability and drop in performance against iterative attacks (PGD and APGD) remain almost the same regardless of the model's size.
\begin{table}[!h]
\setlength{\tabcolsep}{0.5cm}
\centering
\caption{Comparison between models having different scales of LLM but the same vision encoder for \(\epsilon=4/255\). Note: Higher values (\(\uparrow\)) indicate better performance.}
\vspace{-2mm}
\resizebox{\columnwidth}{!}{%
\begin{tabular}{llrrrrrr} \toprule
 &  & \multicolumn{6}{c}{Task} \\
 & \multirow{-2}{*}{Attack} & \multicolumn{1}{c}{COCO} & \multicolumn{1}{c}{Flickr30k} & \multicolumn{1}{c}{OK-VQA} & \multicolumn{1}{c}{TextVQA} & \multicolumn{1}{c}{VizWiz} & \multicolumn{1}{c}{VQAv2} \\ \toprule
 & None & 119.02 & 77.21 & 59.18 & 36.73 & 39.94 & 70.00 \\
 & FGSM & 96.79 & 64.25 & 49.70 & 25.32 & 33.92 & 56.52 \\
 & PGD & 18.54 & 13.57 & 16.26 & 9.44 & 10.67 & 28.60 \\
\multirow{-4}{*}{LLaVA-7B} & APGD & 30.20 & 22.11 & \cellcolor[HTML]{F0DAE3}22.86 & 11.73 & 22.86 & 28.76 \\ \midrule
 & None & 123.71 & 77.63 & 62.86 & 40.04 & 41.19 & 75.39 \\
 & FGSM & \cellcolor[HTML]{E5F3E0}123.71 & \cellcolor[HTML]{E5F3E0}77.63 & \cellcolor[HTML]{E5F3E0}62.86 & \cellcolor[HTML]{E5F3E0}40.04 & \cellcolor[HTML]{E5F3E0}41.19 & \cellcolor[HTML]{E5F3E0}75.39 \\
 & PGD & 18.54 & 13.57 & 16.26 & 9.44 & 10.67 & 28.60 \\
\multirow{-4}{*}{LLaVA-13B} & APGD & 30.20 & 22.11 & 21.30 & 11.73 & 22.86 & 28.76 \\ \bottomrule
\end{tabular}%
}
\label{tab:comparison_llm_size_eps_4}
\end{table}

\begin{table}[!h]
\setlength{\tabcolsep}{0.5cm}
\centering
\caption{Comparison between models having different scales of LLM but the same vision encoder for  \(\epsilon = 16/255\). Note: Higher values (\(\uparrow\)) indicate better performance.}
\vspace{-2mm}
\resizebox{\columnwidth}{!}{%
\begin{tabular}{llrrrrrr} \toprule
 &  & \multicolumn{6}{c}{Task} \\
 & \multirow{-2}{*}{Attack} & \multicolumn{1}{c}{COCO} & \multicolumn{1}{c}{Flickr30k} & \multicolumn{1}{c}{OK-VQA} & \multicolumn{1}{c}{TextVQA} & \multicolumn{1}{c}{VizWiz} & \multicolumn{1}{c}{VQAv2} \\ \toprule
 & None & 119.02 & 77.21 & 59.18 & 36.73 & 39.94 & 70.00 \\
 & FGSM & \cellcolor[HTML]{E5F3E0}101.98 & 56.08 & 49.24 & 22.20 & 37.39 & 58.30 \\
 & PGD & 10.08 & 5.25 & \cellcolor[HTML]{DBECFC}17.48 & 6.70 & 8.52 & 24.39 \\
\multirow{-4}{*}{LLaVA-7B} & APGD & 13.88 & \cellcolor[HTML]{F0DAE3}9.05 & 13.26 & 7.34 & 27.54 & 19.98 \\ \midrule
 & None & 123.71 & 77.63 & 62.86 & 40.04 & 41.19 & 75.39 \\
 & FGSM & 99.83 & \cellcolor[HTML]{E5F3E0}58.40 & \cellcolor[HTML]{E5F3E0}52.90 & \cellcolor[HTML]{E5F3E0}24.85 & \cellcolor[HTML]{E5F3E0}37.89 & \cellcolor[HTML]{E5F3E0}60.98 \\
 & PGD & 10.08 & \cellcolor[HTML]{DBECFC}9.05 & 13.26 & 6.70 & 8.52 & 24.39 \\
\multirow{-4}{*}{LLaVA-13B} & APGD & 13.88 & 5.25 & \cellcolor[HTML]{F0DAE3}17.48 & 7.34 & 27.54 & 19.98 \\ \bottomrule
\end{tabular}%
}
\label{tab:comparison_llm_size_eps_16}
\end{table}

\newpage
\subsection{Ensemble of Vision Encoders}

The observations for both \(\epsilon=4/255\) and \(16/255\) are same as for \(\epsilon=8/255\). Targeting the weakest image encoder is enough to jeopardize the entire system. Conversely, having the strongest vision encoder in the ensemble ensures the best robust performance.
\begin{table}[!h]
\setlength{\tabcolsep}{0.5cm}
\centering
\caption{Comparison between VLMs that have an ensemble of vision encoders. The comparison is made when only the input to the Dino image encoder is perturbed for \(\epsilon = 4/255\). Note: Higher values (\(\uparrow\)) indicate better performance.}

\resizebox{\columnwidth}{!}{%
\begin{tabular}{llrrrrrr} \toprule
 &  & \multicolumn{6}{c}{Task} \\
 & \multirow{-2}{*}{Attack} & \multicolumn{1}{c}{COCO} & \multicolumn{1}{c}{Flickr30k} & \multicolumn{1}{c}{OK-VQA} & \multicolumn{1}{c}{TextVQA} & \multicolumn{1}{c}{VizWiz} & \multicolumn{1}{c}{VQAv2} \\ \toprule
 & None & 113.75 & 74.16 & 58.88 & 15.08 & 39.30 & 74.35 \\
 & FGSM & 99.01 & 57.19 & 47.14 & 8.92 & 38.39 & 58.50 \\
 & PGD & 21.00 & 12.22 & 14.96 & 3.34 & 13.95 & 35.03 \\
\multirow{-4}{*}{DinoCLIP} & APGD & 10.71 & 7.12 & 12.96 & 3.30 & 12.49 & 33.10 \\ \midrule
 & None & 125.94 & 85.44 & 61.12 & 50.52 & 44.27 & 79.39 \\
 & FGSM & \cellcolor[HTML]{E5F3E0}107.87 & \cellcolor[HTML]{E5F3E0}74.10 & \cellcolor[HTML]{E5F3E0}52.92 & \cellcolor[HTML]{E5F3E0}40.36 & \cellcolor[HTML]{E5F3E0}40.77 & \cellcolor[HTML]{E5F3E0}67.24 \\
 & PGD & \cellcolor[HTML]{DBECFC}52.89 & \cellcolor[HTML]{DBECFC}32.14 & \cellcolor[HTML]{DBECFC}23.10 & \cellcolor[HTML]{DBECFC}15.73 & \cellcolor[HTML]{DBECFC}14.78 & \cellcolor[HTML]{DBECFC}42.87 \\
\multirow{-4}{*}{DinoSigLIP} & APGD & \cellcolor[HTML]{F0DAE3}35.34 & \cellcolor[HTML]{F0DAE3}22.86 & \cellcolor[HTML]{F0DAE3}19.18 & \cellcolor[HTML]{F0DAE3}13.83 & \cellcolor[HTML]{F0DAE3}12.96 & \cellcolor[HTML]{F0DAE3}39.58 \\ \bottomrule
\end{tabular}%
}
\label{tab:comparison_ensemble_ve_eps_4}
\end{table}

\begin{table}[!h]
\setlength{\tabcolsep}{0.5cm}
\centering
\caption{Comparison between VLMs that have an ensemble of vision encoders. The comparison is made when only the input to the Dino image encoder is perturbed for \(\epsilon = 16/255\). Note: Higher values (\(\uparrow\)) indicate better performance.}

\resizebox{\columnwidth}{!}{%
\begin{tabular}{llrrrrrr} \toprule
 &  & \multicolumn{6}{c}{Task} \\
 & \multirow{-2}{*}{Attack} & \multicolumn{1}{c}{COCO} & \multicolumn{1}{c}{Flickr30k} & \multicolumn{1}{c}{OK-VQA} & \multicolumn{1}{c}{TextVQA} & \multicolumn{1}{c}{VizWiz} & \multicolumn{1}{c}{VQAv2} \\ \toprule
 & None & 113.75 & 74.16 & 58.88 & 15.08 & 39.30 & 74.35 \\
 & FGSM & 103.37 & 57.80 & 48.38 & 9.29 & 40.30 & 58.87 \\
 & PGD & 8.14 & 5.73 & 11.22 & 2.86 & 12.12 & 32.38 \\
\multirow{-4}{*}{DinoCLIP} & APGD & 3.11 & 2.37 & 8.94 & 2.57 & \cellcolor[HTML]{F0DAE3}12.63 & 27.47 \\ \midrule
 & None & 125.94 & 85.44 & 61.12 & 50.52 & 44.27 & 79.39 \\
 & FGSM & \cellcolor[HTML]{E5F3E0}111.52 & \cellcolor[HTML]{E5F3E0}74.29 & \cellcolor[HTML]{E5F3E0}54.76 & \cellcolor[HTML]{E5F3E0}42.78 & \cellcolor[HTML]{E5F3E0}42.74 & \cellcolor[HTML]{E5F3E0}68.40 \\
 & PGD & \cellcolor[HTML]{DBECFC}31.29 & \cellcolor[HTML]{DBECFC}17.58 & \cellcolor[HTML]{DBECFC}18.52 & \cellcolor[HTML]{DBECFC}12.72 & \cellcolor[HTML]{DBECFC}12.61 & \cellcolor[HTML]{DBECFC}39.06 \\
\multirow{-4}{*}{DinoSigLIP} & APGD & \cellcolor[HTML]{F0DAE3}17.77 & \cellcolor[HTML]{F0DAE3}10.57 & \cellcolor[HTML]{F0DAE3}14.86 & \cellcolor[HTML]{F0DAE3}10.94 & 12.38 & \cellcolor[HTML]{F0DAE3}35.66 \\ \bottomrule
\end{tabular}%
}
\label{tab:comparison_ensemble_ve_eps_16}
\end{table}


\section{Prompt Formatting}
\label{appendix:prompt_formatting}
\begin{table}[!h]
\centering
\resizebox{\textwidth}{!}{%
\begin{tabular}{ll} \toprule
Task & \multicolumn{1}{c} {Instruction} \\ \midrule
Rephrase & \begin{tabular}[c]{@{}l@{}}\texttt{You will be given a question. Your task is to rephrase the question so that it}\\ \texttt{is semantically similar to the original question and will have the same answer}\\ \texttt{as the original question.}\end{tabular} \\ \midrule
Expand & \begin{tabular}[c]{@{}l@{}}\texttt{You will be given a short question. Your task is to generate a longer question so that it}\\ \texttt{is semantically similar to the original question and will have the same answer} \\ \texttt{as the original question.}\end{tabular}\\ \midrule
AC & \begin{tabular}[c]{@{}l@{}}\texttt{You will be given a question. However, the image associated with the question will be} \\\texttt{adversarially perturbed. Your task is to generate a longer question so that it is semantically}\\ \texttt{similar to the original question and will have the same answer as the original question.} \end{tabular} \\
\midrule
AP & \begin{tabular}[c]{@{}l@{}}\texttt{You will be given a question. However, the image associated with the question could be} \\\texttt{adversarially perturbed. Your task is to generate a longer question so that it is semantically}\\ \texttt{similar to the original question and will have the same answer as the original question.} \end{tabular} \\
\bottomrule
\end{tabular}%
}
\caption{Instructions used to obtain the modified questions for VQA.}
\label{tab:vqa-prompts}
\end{table}

\begin{table}[!h]
\centering
\resizebox{\textwidth}{!}{%
\begin{tabular}{lll} \toprule
 \multicolumn{1}{l}{Prompt Type} & \multicolumn{1}{c}{Prompt} \\ \toprule
  Original & \texttt{Provide a short caption for this image.} \\ \midrule
 AC & \begin{tabular}[c]{@{}l@{}}\texttt{Consider the given image being adversarially perturbed. Provide a short caption for this image.}\end{tabular} \\  \midrule
 AP & \begin{tabular}[c]{@{}l@{}}\texttt{Given image could be adversarially perturbed. Provide a short caption for this image.}\end{tabular} \\ \midrule
 Random sent. & \begin{tabular}[c]{@{}l@{}}\texttt{Clouds drift quietly over the ancient, forgotten city. Provide a short caption for this image.}\end{tabular} \\ \midrule
  Random str. & \begin{tabular}[c]{@{}l@{}}\texttt{ryFo8ZVcyNMtLgryNOg64UTjySyEb79e5aq6IJxGuz0GzWNtoz. Provide a short caption for this image.}\end{tabular}  &\\ \bottomrule
\end{tabular}%
}
\caption{Various types of prompts tested for image captioning.}
\label{tab:prompts-captioning}
\end{table}



\newpage

\section{Prompt Formatting Results}
\label{appendix:prompt_formatting_results}
\begin{table}[!h]
\centering
\caption{Performance of LLaVA models on image captioning (COCO) when adversarially perturbed images (using \(\epsilon=8/255\)) are provided along with different types of prompts. Note: Higher values (\(\uparrow\)) indicate better performance.}

\begin{tabular}{llrrrr} \toprule
\textbf{} & \multicolumn{1}{c}{Prompt} & \multicolumn{1}{l}{FGSM} & \multicolumn{1}{l}{PGD} & \multicolumn{1}{l}{APGD} & \multicolumn{1}{l}{Clean} \\ \toprule
\multirow{6}{*}{LLaVA 7B} & Original & 95.55 & 22.84 & 12.54 & 119.02 \\
 & AC & 63.86 & 60.01 & 54.05 & 64.11 \\
 & AP & 105.41 & 101.61 & 91.46 & 112.78 \\
 & Random str & 108.23 & 105.35 & 94.61 & 120.90 \\
 & Random sent & 101.12 & 97.11 & 88.45 & 108.00 \\ \cmidrule{2-6}
 & Clean Acc & \multicolumn{4}{r}{119.02} \\ \midrule
\multirow{6}{*}{LLaVA 13B} & Original & 106.40 & 14.60 & 6.98 & 123.71 \\
 & AC & 113.77 & 106.40 & 114.65 & 122.10 \\
 & AP & 114.48 & 108.83 & 113.54 & 125.28 \\
 & Random str & 110.74 & 105.15 & 111.69 & 120.49 \\
 & Random sent & 113.29 & 106.71 & 111.13 & 120.72 \\ \cmidrule{2-6}
 & Clean acc & \multicolumn{4}{r}{123.71} \\ \bottomrule
\end{tabular}%
\label{tab:captioning-prompting}
\end{table}



\begin{table}[!h]
\centering
\caption{Performance of LLaVA models on VQAv2 when adversarially perturbed images (using \(\epsilon=8/255\)) are provided along with questions generated using different types of prompts. Note: Higher values (\(\uparrow\)) indicate better performance.}

\begin{tabular}{llrrrr} \toprule
\textbf{} & \multicolumn{1}{c}{Prompt} & \multicolumn{1}{c}{FGSM} & \multicolumn{1}{c}{PGD} & \multicolumn{1}{c}{APGD} & \multicolumn{1}{c}{Clean} \\ \toprule
\multirow{6}{*}{LLaVA 7B} & Original & 57.19 & 27.07 & 26.73 & 70.00 \\
 & Rephrase & 59.01 & 58.03 & 48.84 & 68.30 \\
 & AC & 60.21 & 58.82 & 50.68 & 69.99 \\
 & AP & 60.13 & 58.81 & 49.95 & 69.78 \\
 & Expand & 48.59 & 48.54 & 42.24 & 57.14 \\ \cmidrule{2-6}
 & Clean Acc & \multicolumn{4}{r}{70.00} \\ \midrule
\multirow{6}{*}{LLaVA 13B} & Original & 62.49 & 34.49 & 33.56 & 75.39 \\
 & Rephrase & 59.01 & 60.05 & 54.77 & 71.02 \\
 & AC & 51.38 & 61.49 & 55.56 & 72.00 \\
 & AP & 63.59 & 61.29 & 63.2 & 71.79 \\
 & Expand & 53.03 & 50.03 & 45.93 & 58.59 \\ \cmidrule{2-6}
 & Clean Acc & \multicolumn{4}{r}{75.39} \\ \bottomrule
\end{tabular}%
\label{tab:vqa-prompting}
\end{table}

\end{document}